\documentclass[abstract, final]{nldl}

\paperID{17}

\vol{V}

\usepackage{mathtools}
\usepackage{amsmath}
\usepackage{amsmath}   
\usepackage{amssymb}   
\usepackage{amsfonts}  
\usepackage[inline]{enumitem}
\usepackage{graphicx}

\usepackage{booktabs}

\usepackage{enumitem}

\usepackage{algorithm}
\usepackage{algorithmic}

\usepackage{listings}
\usepackage{hyperref}
\usepackage{url}
\hypersetup{
  pdfusetitle,
  colorlinks,
  linkcolor = BrickRed,
  citecolor = NavyBlue,
  urlcolor  = Magenta!80!black,
}

\title{Is it the model or the metric - On robustness measures of deeplearning models \thanks{Extended abstract at Northern Lights Deep Learning (NLDL) Conference 2025}}
\author[1]{Zhijin Lyu}
\author[2]{Yutong Jin}
\author[3]{Sneha Das \thanks{Corresponding Author. sned@dtu.dk}}

\affil[1]{Dept. Electrical and Photonics Engineering, Technical University of Denmark, Lyngby, Denmark}
\affil[2]{Dept. Engineering, King's College London, London, UK}
\affil[3]{Dept. Applied Mathematics and Computer Science, Technical University of Denmark, Lyngby, Denmark}
\date{}

\begin{document}
\maketitle

\begin{abstract}
Determining the robustness of deep learning models is an established and ongoing challenge within automated decision-making systems. With the advent and success of techniques that enable advanced deep learning (DL), these models are being used in widespread applications, including high-stake ones like healthcare, education, border-control. Therefore, it is critical to understand the limitations of these models and predict their regions of failures, in order to create the necessary guardrails for their successful {\it and} safe deployment. In this work, we revisit robustness, specifically investigating the sufficiency of {\it robust accuracy} (RA), within the context of deepfake detection. We present robust ratio (RR) as a complementary metric, that can quantify the changes to the normalized or probability outcomes under input perturbation. We present a comparison of RA and RR and demonstrate that despite similar RA between models, the models show varying RR under different tolerance (perturbation) levels.
\end{abstract}
\vspace{-.5cm}
\section{Introduction}
Adversarial-robustness and its measurement in deep learning models is a long established and active research domain \cite{goodfellow2014explaining,zhao2024noise, chaudhury2024adversarial, theodoropoulos2024federated, deng2024noise,liu2024understanding,shah2024fixed}. Designing adversarial attacks and developing models that can resist these adversarial attacks have been a primary focus in domain. However, there is still a general lack in consensus and understanding of what comprises as a standard robustness-test of model. There is growing call to develop more standardized measures to evaluate the vulnerability and robustness of deep-models before deployment \cite{akhtar2024video}. Such measures are integral to broader artificial intelligence (AI) safety objectives, ensuring that models function reliably and ethically, even in complex and high-stakes applications.
Therefore, in line with recent regulatory requirements, we need to both develop measures and a consensus in the community on the standard definitions and measure of robustness~\cite{AI_act, speinshart}.   

In this work, we investigate {\it robust accuracy (RA)}, which is a widely used metric to measure robustness of models in adversarial settings, within the context of deepfake detection~\cite{rana2022deepfake}. RA is similar to metrics like adversarial accuracy, commonly used in literature~\cite{guo2023comprehensive}.
RA measures the percentage of correctly classified adversarial examples, quantifying only when the model misclassifies. 
This, however does not account for the changes in model outcomes under perturbations, which are not sufficiently large to change the class outcome. We argue that the output changes under input perturbations are also crucial to evaluate the models against. 
Employing the definition of robustness from {\it formal verification}~\cite{bensalem2024bridging}, we present {\it robust ratio (RR)}, which captures the proportion of examples where arbitrary input perturbations result in equivalent changes in normalized continuous (or probability) outcome of the model. 
This makes RR particularly useful in applications like healthcare, where it is necessary to quantify the optimum region of model-operation. Alternative metrics like robustness-score and average-confidence-of-class are similar to RR in providing a more granular evaluation of model behavior under adversarial perturbations, and we will explore the interaction between these metrics in future work~\cite{laugros2019adversarial, guo2023comprehensive}.
\vspace{-.6cm}
\section{Robustness Metric}
In the following sections, we formally describe RA~(Eq.~\ref{eq:1}) and RR~(Eqs.~\ref{eq:2} \&~\ref{eq:3}).\\
\noindent
{\bf Robust accuracy} 
computes the proportion of predictions $\hat{y}$ that remain correct under an adversarial perturbation $\delta$ as follows:
\begin{equation}\label{eq:1}
\text{Robust Accuracy} \triangleq \frac{\sum_{i=1}^{N} \mathbb{1} \left( \hat{y}_i = y_i \right)}{N} 
\end{equation}
where $N$ is the total number of samples, $\hat{y}_i$ represents the model’s predicted label after perturbation, $y_i$ is the ground truth label, and $\mathbb{1}$ here is the indicator function. As can be seen observed from Eq.~\ref{eq:1}, while RA quantifies whether the model's output class is accurate, it does not account for how the output probabilities changes as a result of the input perturbations.

\begin{figure*}[!htb]
  \centering
  \begin{subfigure}{0.48\linewidth}
    \centering
    \includegraphics[width=\linewidth]{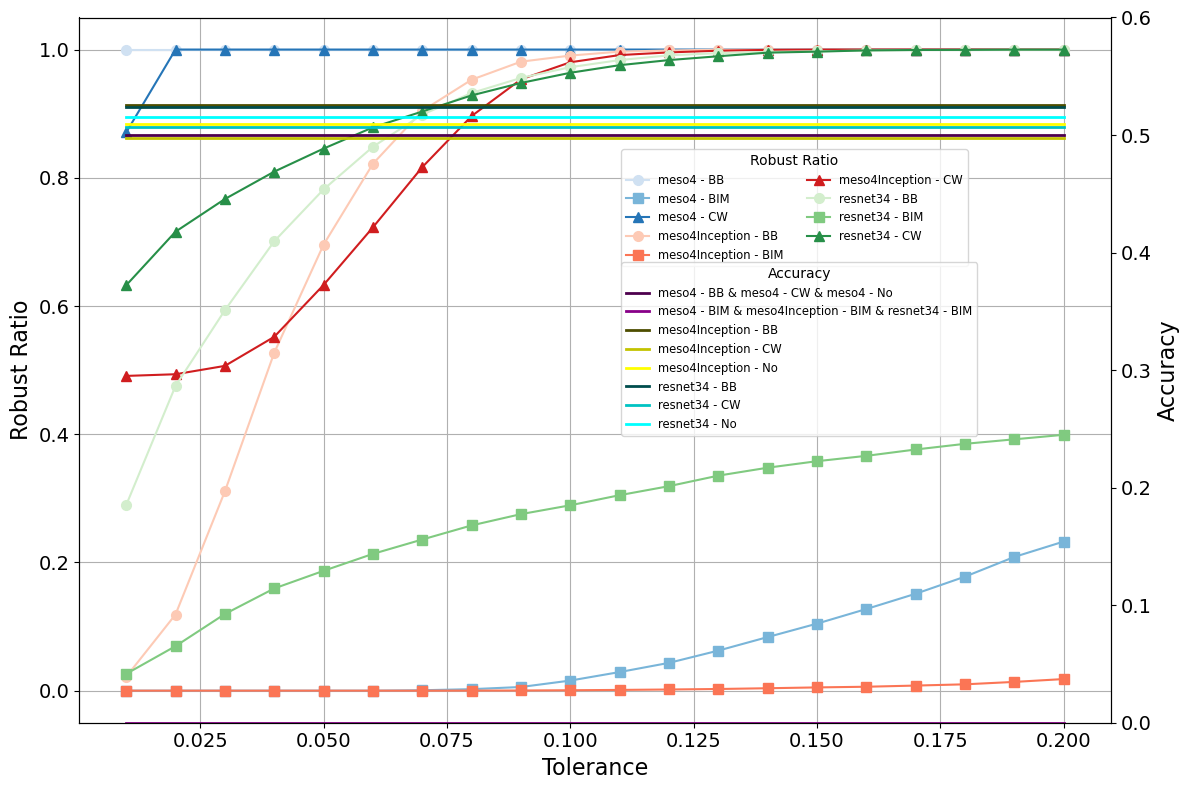}
    \caption{Image dataset}
    \label{fig:image_tolerance}
  \end{subfigure}
  \hfill
  \begin{subfigure}{0.48\linewidth}
    \centering
    \includegraphics[width=\linewidth]{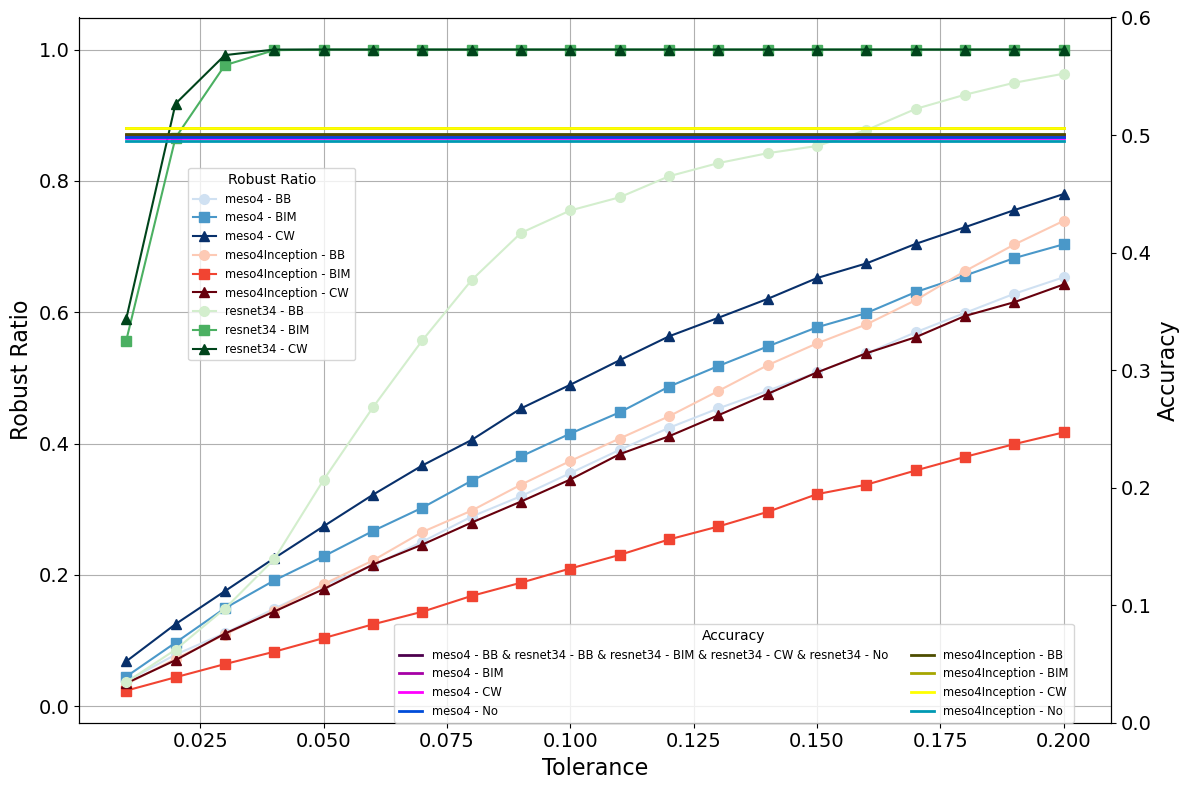}
    \caption{Video dataset}
    \label{fig:video_tolerance}
  \end{subfigure}
  \caption{Comparison of a) Robust ratio vs. Tolerance on left-y-axis, and b) robust accuracy on right-y-axis.}
  \label{fig:combined}
  \vspace{-0.5cm}
\end{figure*}
\noindent
{\bf Robust Ratio:} Using the definition of robustness introduced in the MLS$^{2}$ framework \cite{bensalem2024bridging} in Eq.~\ref{eq:2}, we obtain the numerator for RR~(Eq.~\ref{eq:3}). It captures the proportion of samples where small perturbations in the input, $\delta$, result in a {\it bounded} change in the normalized output or the probability. The robustness of a model $\phi^{i}_{\text{rob}}$, with weight matrix $\mathbf{w_{i}}$ over input $\mathbf{x_{i}}$, holds when:
\begin{align}\label{eq:2}
\phi^{i}_{\text{rob}}(\mathbf{w_{i}}, \mathbf{x_{i}}) \triangleq \forall \mathbf{\delta} : \|\mathbf{\delta}\|_{2} \leq  \epsilon \Rightarrow \notag \\
\quad \left| P(Y \mid \mathbf{x_{i}} + \mathbf{\delta}, \mathbf{w_{i}})(\hat{y}) - P(Y \mid \mathbf{x_{i}}, \mathbf{w_{i}})(\hat{y}) \right| \leq \text{b}
\end{align}
where $\|\mathbf{\delta}\|_2$ is the $\L_{2}$ norm of the perturbation, $\mathbf{\delta} \in \mathbb{R}^{n}$ is the perturbation vector, and $\epsilon$ is the threshold for perturbation, $b$ is the bound on the continuous (normalized or probability) outcome of the model. The key difference is that RR allows the quantification of how stable the continuous output remains within a desired perturbation range, even if the final predicted label does not change.
Further, robust ratio is computed as follows:
\begin{equation}\label{eq:3}
Robust Ratio \triangleq \frac{\sum_{i=1}^{N} \phi^{i}_{\text{rob}}(\mathbf{w_i}, \mathbf{x_i})}{N}
\end{equation}
where $N$ is the total number of samples in the dataset. This metric reflects the overall robustness performance by assessing the proportion of samples that satisfy local robustness conditions under small adversarial perturbations. A higher RR indicates that the model is more resilient to small input perturbations for the majority of samples.

\section{Experimental Setup}
{\bf Datasets:} For this study, we compiled image and video samples using the following datasets: UADFV \cite{yang2019exposing}, FF++ \cite{rossler2019faceforensics++}, and Celeb-DF-v2 \cite{li2019celeb}. The image dataset includes 1280 images for each of following six demographic groups: \begin{enumerate*}\item African-American-male, \item African-American-female, \item Caucasian-male, \item Caucasian-female, \item Asian-male, and \item Asian-female.\end{enumerate*} Each group has 640 real and 640 fake images, resulting in a total of 7680 images. Additionally, the video dataset consists of 5 videos, each 2 seconds in length for the above demographic groups. As preprocessing, we applied face detection and image resizing to ensure consistent data formatting and eliminate additional sources of variance between the samples.\\
\noindent
{\bf Models:} We investigate the following deepfake detection models: Meso4~\cite{afchar2018mesonet}, Meso4Inception~\cite{afchar2018mesonet}, and ResNet34~\cite{koonce2021resnet}. These models were selected based on their performance during initial tests and their efficiency in terms of energy consumption. \\
\noindent
{\bf Adversarial Attacks:} We tested the robustness of the models using the commonly used adversarial attack methods: fast gradient sign method (FGSM)~\cite{goodfellow2014explaining}, projected gradient descent (PGD)~\cite{madry2017towards} and Carlini and Wagner (CW)~\cite{carlini2017towards}. \\
\noindent
{\bf Tolerance Settings:} To measure the models' response to different perturbation magnitudes, we compute RR at different tolerance (proportional to the bound) levels; tolerance $\delta = \{0, .01, .02,..., 0.2\}.$ 
\vspace{-.0cm}
\section{Results and discussion}
The results in terms of the RA and RR over the three models and attacks are presented in Fig.~\ref{fig:combined}. 
We observe that RA, function of only the attack type, shows similar performances for all models and over all attack types, for both images and videos. In contrast, RR shows that the performance of models is largely influenced by the tolerance and bound values, highlighting that models have different optimum regions of operation. Defining these optimum regions can aid in choosing models suitable to the application scenario (eg: high-noise clinical data versus carefully curated developmental benchmark). 

Furthermore, RR exhibits a more linear relationship with tolerance for video datasets compared to image datasets. 
We hypothesize that this difference may be attributed to variations in the implementation of adversarial attacks across the two types of datasets. Specifically, we suspect that in video datasets, the adversarial perturbations are applied more progressively or iteratively across frames, leading to a gradual accumulation of the perturbation over time. This contrasts with image datasets, where perturbations may be applied more directly to the entire frame, leading to more abrupt effects. In other words, the temporal dimension in video datasets may result in smoother and more cumulative changes, explaining the linear relationship between RR and tolerance, and a relatively lower RR as well.

\printbibliography

@article{goodfellow2014explaining,
  title={Explaining and harnessing adversarial examples},
  author={Goodfellow, Ian J and Shlens, Jonathon and Szegedy, Christian},
  journal={arXiv preprint arXiv:1412.6572},
  year={2014}
}

@inproceedings{laugros2019adversarial,
  title={Are adversarial robustness and common perturbation robustness independant attributes?},
  author={Laugros, Alfred and Caplier, Alice and Ospici, Matthieu},
  booktitle={Proceedings of the IEEE/CVF International Conference on Computer Vision Workshops},
  pages={0--0},
  year={2019}
}

@inproceedings{zhao2024noise,
  title={Noise-BERT: A Unified Perturbation-Robust Framework with Noise Alignment Pre-Training for Noisy Slot Filling Task},
  author={Zhao, Jinxu and Dong, Guanting and Qiu, Yueyan and Hui, Tingfeng and Song, Xiaoshuai and Guo, Daichi and Xu, Weiran},
  booktitle={ICASSP 2024-2024 IEEE International Conference on Acoustics, Speech and Signal Processing (ICASSP)},
  pages={6150--6154},
  year={2024},
  organization={IEEE}
}

@inproceedings{chaudhury2024adversarial,
  title={Adversarial Robustness of Convolutional Models Learned in the Frequency Domain},
  author={Chaudhury, Subhajit and Yamasaki, Toshihiko},
  booktitle={ICASSP 2024-2024 IEEE International Conference on Acoustics, Speech and Signal Processing (ICASSP)},
  pages={7455--7459},
  year={2024},
  organization={IEEE}
}

@inproceedings{theodoropoulos2024federated,
  title={Federated Learning under Restricted user Availability},
  author={Theodoropoulos, Periklis and Nikolakakis, Konstantinos E and Kalogerias, Dionysis},
  booktitle={ICASSP 2024-2024 IEEE International Conference on Acoustics, Speech and Signal Processing (ICASSP)},
  pages={7055--7059},
  year={2024},
  organization={IEEE}
}

@inproceedings{deng2024noise,
  title={Noise-Resistant Graph Neural Network for Node Classification},
  author={Deng, Zichao and Yu, Han},
  booktitle={ICASSP 2024-2024 IEEE International Conference on Acoustics, Speech and Signal Processing (ICASSP)},
  pages={7560--7564},
  year={2024},
  organization={IEEE}
}

@inproceedings{liu2024understanding,
  title={Understanding Data Augmentation From A Robustness Perspective},
  author={Liu, Zhendong and Zhang, Jie and He, Qiangqiang and Wang, Chongjun},
  booktitle={ICASSP 2024-2024 IEEE International Conference on Acoustics, Speech and Signal Processing (ICASSP)},
  pages={6760--6764},
  year={2024},
  organization={IEEE}
}

@inproceedings{shah2024fixed,
  title={Fixed Inter-Neuron Covariability Induces Adversarial Robustness},
  author={Shah, Muhammad A and Raj, Bhiksha},
  booktitle={ICASSP 2024-2024 IEEE International Conference on Acoustics, Speech and Signal Processing (ICASSP)},
  pages={7005--7009},
  year={2024},
  organization={IEEE}
}

@article{bensalem2024bridging,
  title={Bridging formal methods and machine learning with model checking and global optimisation},
  author={Bensalem, Saddek and Huang, Xiaowei and Ruan, Wenjie and Tang, Qiyi and Wu, Changshun and Zhao, Xingyu},
  journal={Journal of Logical and Algebraic Methods in Programming},
  volume={137},
  pages={100941},
  year={2024},
  publisher={Elsevier}
}

@article{akhtar2024video,
  title={Video and audio deepfake datasets and open issues in deepfake technology: being ahead of the curve},
  author={Akhtar, Zahid and Pendyala, Thanvi Lahari and Athmakuri, Virinchi Sai},
  journal={Forensic Sciences},
  volume={4},
  number={3},
  pages={289--377},
  year={2024},
  publisher={MDPI}
}

@misc{AI_act,
  author = {European Union},
  title = {Regulation (EU) 2024/1689 of the European Parliament and of the Council on harmonised rules on Artificial Intelligence (AI Act)},
  howpublished = "https://eur-lex.europa.eu/legal-content/EN/TXT/?uri=CELEX:32024R1689",
  year = {2024}, 
  note = "[Online; accessed September-2024]"
}

@techreport{speinshart,
 title={The Speinshart Recommendations on Generative AI and the EU AI Act},
  author={D Ahern and I Bratko and S Das and S Delacroix and L Kastner and L Fetic and T Klein and D Lewis and A Mazumder and P Molnar and L Nieper and AA Faisal},
     url = {https://doi.org/10.25561/108241},
     year = {2023},
     institution = {Imperial College London},
     month = {09}
}

@article{rana2022deepfake,
  title={Deepfake detection: A systematic literature review},
  author={Rana, Md Shohel and Nobi, Mohammad Nur and Murali, Beddhu and Sung, Andrew H},
  journal={IEEE access},
  volume={10},
  pages={25494--25513},
  year={2022},
  publisher={IEEE}
}

@article{guo2023comprehensive,
  title={A comprehensive evaluation framework for deep model robustness},
  author={Guo, Jun and Bao, Wei and Wang, Jiakai and Ma, Yuqing and Gao, Xinghai and Xiao, Gang and Liu, Aishan and Dong, Jian and Liu, Xianglong and Wu, Wenjun},
  journal={Pattern Recognition},
  volume={137},
  pages={109308},
  year={2023},
  publisher={Elsevier}
}

@inproceedings{rossler2019faceforensics++,
  title={Faceforensics++: Learning to detect manipulated facial images},
  author={Rossler, Andreas and Cozzolino, Davide and Verdoliva, Luisa and Riess, Christian and Thies, Justus and Nie{\ss}ner, Matthias},
  booktitle={Proceedings of the IEEE/CVF international conference on computer vision},
  pages={1--11},
  year={2019}
}

@article{li2019celeb,
  title={Celeb-df (v2): a new dataset for deepfake forensics [j]},
  author={Li, Yuezun and Yang, Xin and Sun, Pu and Qi, Honggang and Lyu, Siwei},
  journal={arXiv preprint arXiv},
  year={2019}
}

@inproceedings{yang2019exposing,
  title={Exposing deep fakes using inconsistent head poses},
  author={Yang, Xin and Li, Yuezun and Lyu, Siwei},
  booktitle={ICASSP 2019-2019 IEEE International Conference on Acoustics, Speech and Signal Processing (ICASSP)},
  pages={8261--8265},
  year={2019},
  organization={IEEE}
}

@inproceedings{afchar2018mesonet,
  title={Mesonet: a compact facial video forgery detection network},
  author={Afchar, Darius and Nozick, Vincent and Yamagishi, Junichi and Echizen, Isao},
  booktitle={2018 IEEE international workshop on information forensics and security (WIFS)},
  pages={1--7},
  year={2018},
  organization={IEEE}
}

@article{koonce2021resnet,
  title={ResNet 34},
  author={Koonce, Brett and Koonce, Brett},
  journal={Convolutional neural networks with swift for tensorflow: image recognition and dataset categorization},
  pages={51--61},
  year={2021},
  publisher={Springer}
}

@article{madry2017towards,
  title={Towards deep learning models resistant to adversarial attacks},
  author={Madry, Aleksander and Makelov, Aleksandar and Schmidt, Ludwig and Tsipras, Dimitris and Vladu, Adrian},
  journal={arXiv preprint arXiv:1706.06083},
  year={2017}
}

@inproceedings{carlini2017towards,
  title={Towards evaluating the robustness of neural networks},
  author={Carlini, Nicholas and Wagner, David},
  booktitle={2017 ieee symposium on security and privacy (sp)},
  pages={39--57},
  year={2017},
  organization={Ieee}
}
\end{document}